\documentclass[sigconf]{acmart}

\usepackage{balance}

\AtBeginDocument{%
  \providecommand\BibTeX{{%
    \normalfont B\kern-0.5em{\scshape i\kern-0.25em b}\kern-0.8em\TeX}}}

\copyrightyear{2022} 
\acmYear{2022} 
\setcopyright{rightsretained} 

\acmConference[SIGIR '22]{Proceedings of the 45th International ACM SIGIR Conference on Research and Development in Information Retrieval}{July 11--15, 2022}{Madrid, Spain.}
\acmBooktitle{Proceedings of the 45th Int'l ACM SIGIR Conference on Research and Development in Information Retrieval (SIGIR '22), July 11--15, 2022, Madrid, Spain}
\acmDOI{10.1145/3477495.3531979}
\acmISBN{978-1-4503-8732-3/22/07}

\usepackage{etoolbox}
\makeatletter
\patchcmd{\maketitle}{\@copyrightpermission}{
\begin{minipage}{0.3\columnwidth}
\href{http://creativecommons.org/licenses/by/4.0/}{\includegraphics[width=0.90\textwidth]{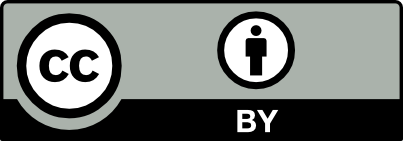}}
\end{minipage}\hfill
\begin{minipage}{0.7\columnwidth}
\href{http://creativecommons.org/licenses/by/4.0/}{This work is licensed under a Creative Commons Attribution International 4.0 License.}
\end{minipage}

\vspace{5pt}
}{}{}

\makeatother

\begin{document}
\fancyhead{}

\title{Few-Shot Stance Detection via Target-Aware Prompt Distillation}

\author{Yan Jiang}
\affiliation{%
\institution{\textsuperscript{1}Data Intelligence System Research Center, Institute of Computing Technology, Chinese Academy of Sciences}
\institution{\textsuperscript{2}University of Chinese Academy of Sciences}
\city{Beijing}
\country{China}
}
 \email{jiangyan18s@ict.ac.cn}

\author{Jinhua Gao}
\authornote{Jinhua Gao is the corresponding author.}
\affiliation{
\institution{Data Intelligence System Research Center, Institute of Computing Technology, Chinese Academy of Sciences}
\city{Beijing}
\country{China}
}
 
\email{gaojinhua@ict.ac.cn}
 
\author{Huawei Shen}
\affiliation{
\institution{\textsuperscript{1}Data Intelligence System Research Center, Institute of Computing Technology, Chinese Academy of Sciences}
\institution{\textsuperscript{2}University of Chinese Academy of Sciences}
\city{Beijing}
\country{China}
}
\email{shenhuawei@ict.ac.cn}
 
\author{Xueqi Cheng}
\affiliation{
\institution{\textsuperscript{1}CAS Key Lab of Network Data Science and Technology, Institute of Computing Technology, Chinese Academy of Sciences}
\institution{\textsuperscript{2}University of Chinese Academy of Sciences}
\city{Beijing}
\country{China}
}
\email{cxq@ict.ac.cn}

\renewcommand{\shortauthors}{Yan Jiang, Jinhua Gao, Huawei Shen and Xueqi Cheng}

\begin{abstract}
Stance detection aims to identify whether the author of a text is in favor of, against, or neutral to a given target. The main challenge of this task comes two-fold: few-shot learning resulting from the varying targets and the lack of contextual information of the targets. Existing works mainly focus on solving the second issue by designing attention-based models or introducing noisy external knowledge, while the first issue remains under-explored. In this paper, inspired by the potential capability of pre-trained language models (PLMs) serving as knowledge bases and few-shot learners, we propose to introduce prompt-based fine-tuning for stance detection. PLMs can provide essential contextual information for the targets and enable few-shot learning via prompts. Considering the crucial role of the target in stance detection task, we design target-aware prompts and propose a novel verbalizer. Instead of mapping each label to a concrete word, our verbalizer maps each label to a vector and picks the label that best captures the correlation between the stance and the target. Moreover, to alleviate the possible defect of dealing with varying targets with a single hand-crafted prompt, we propose to distill the information learned from multiple prompts. Experimental results show the superior performance of our proposed model in both full-data and few-shot scenarios.

\end{abstract}



\maketitle

\section{Introduction}
Stance detection aims to identify whether the author of a text is in favor of, against, or neutral to a given target (\textit{e.g.}, an event, a person, or a claim). This task originates from mining online forums and political debates \cite{thomas2006get}, and has increasingly focused on detecting user stance from social texts in recent years, which is mainly boosted by the tweet stance detection competition and datasets released in SemEval-2016 \cite{mohammad2016semeval}. Stance detection is closely related to sentiment analysis with certain discrepancies. Compared to sentiment analysis, the introduction of the targets makes stance detection more challenging yet practical. A wide range of applications can benefit from stance detection, \textit{e.g.}, information retrieval \cite{sen2018stance}, prediction of election/referendum results \cite{lai2018stance}, rumour classification \cite{zubiaga2018detection}, and fake news detection \cite{lazer2018science}.

As mentioned above, the main challenge of stance detection stems from the introduction of the targets. Firstly, the targets are always time-varying, especially when conducting opinion polling on social media. It is impractical to provide sufficient-scale labeled data for each target, requiring that stance detection methods should be capable of few-shot learning, \textit{i.e.}, working with only a few labeled samples. Secondly, unlike sentiment analysis in which the aspects or targets are explicitly presented, the targets for stance detection may not appear in the text, resulting in the lack of contextual information for understanding the targets. Moreover, the target for stance detection can refer to a claim instead of an entity, \textit{e.g.}, ``Climate Change is a Real Concern'' in SemEval-2016, which further multiplies the challenge. 

Existing works mainly focus on solving the second challenge, while the first one remains under-explored. To capture the contextual information of the targets, most works leverage the attention mechanism to connect targets with their respective contexts. The text and the target are firstly encoded using sequential models like LSTM \cite{rajendran2018stance} or TextCNN \cite{hercig2017detecting}, and then fed into the attention layer to obtain a contextualized representation of the target for inference. However, the contextual information inhabited in the raw texts is usually insufficient. Another line of works introduce external knowledge to address this issue, \textit{e.g.}, mapping words to the semantic lexicon and the emotional lexicon \cite{zhang2020enhancing}, linking words to the entities in commonsense knowledge bases \cite{du2020commonsense}, and aligning words with Wiki concepts \cite{hanawa2019stance}. Though sounding feasible, these works have barely achieved superior performance. The noisy schema linking process hinders their effectiveness, either failing to map words to external information or creating an inaccurate linking. As for few-shot stance detection, existing works typically adopt the cross-target manner, \textit{i.e.}, training on one set of targets and using the trained model to infer the stance on unseen targets \cite{hardalov2021cross,xu2020dan}. However, the expressions and topics are quite different among various targets, limiting the transferability of stance detection models.

Recently, prompt-based fine-tuning on pre-trained language models (PLMs) has proved to be of great potential for few-shot learning \cite{schick2021exploiting,gao2020making,liu2021gpt}. In prompt-based fine-tuning, downstream tasks are first formulated as language modeling problems via prompts, \textit{e.g.}, ``\textit{[sentence]. It was [mask]}'' for sentiment analysis task. Verbalizers then map task labels into concrete words, \textit{e.g.}, ``great'' for positive sentiment and ``terrible'' for the negative. During training and inference, the prompt fills the input sentence into the respective slot and is later fed into the PLM. The label is determined by which verbalized word possesses a higher prediction probability for the \textit{[mask]} token in the prompt. Such a paradigm can better align the downstream task with the pre-training task of the PLM, allowing more comprehensive exploitation of the knowledge implicitly inhabited in the PLM. Moreover, PLMs have proved to store large amounts of relational world knowledge similar to large knowledge graphs in the pre-training stage and prompts can help extract such knowledge \cite{petroni2019language}. This may also benefit capturing the contextual information of the targets for stance detection.

Inspired by the potential capability of PLMs serving as few-shot learners and knowledge bases, we propose to introduce prompt-based fine-tuning for stance detection. However, adapting this framework to stance detection is non-trivial. Firstly, existing verbalizers map each label to the same unique token for all the samples, while the stance label is closely related to the target in stance detection. It is inadequate to share the same set of verbalized label tokens among different targets. Secondly, the design of the prompts is vital yet demanding. Using a single prompt for various targets seems too idealistic. Considering the crucial role of the target in stance detection task, we design target-aware prompts and propose novel verbalizers. Instead of mapping each label to a concrete word, our verbalizer maps each label to a vector. The stance of the text is assigned to the label that best fits the relation ``[target] + [label] $\rightarrow$ [mask]'' where [$\cdot$] denotes the vector representation of the corresponding part. Such a design can fully explore the correlation between the target and the stance. Moreover, to alleviate the possible defect of dealing with varying targets with a single hand-crafted prompt, we incorporate multiple prompts and propose a mutual-distillation mechanism to learn from diverse prompts. We evaluate our model on two datasets and experimental results show the superior performance of our proposed model in both full-data and few-shot scenarios.

Our work makes the following contributions:

\begin{itemize}
\item We propose a prompt-based fine-tuning framework for few-shot stance detection which is under-explored in related research.
\item We design a novel verbalizer to better capture the correlation between the target and the stance, and propose a mutual-distillation mechanism to learn from diverse prompts for varying targets.
\item Extensive experimental results on two datasets show the superior performance of our proposed model in both full-data and few-shot scenarios. Further analysis demonstrates the potential transferability of our model for cross-target stance detection \footnote{Our code and data are available at \url{https://github.com/jyjulyseven/TAPD}.}. 
\end{itemize}

\section{Related Work}
In this section, we introduce works related to stance detection and prompt-based learning. First, we conduct an in-depth review of the existing approaches to identify and classify text stance in Section \ref{section:stancedetection}. Next, we present an overview of recent work done in prompt-based learning in Section \ref{section:prompt-basedlearning}. Last, we introduce works about prompt-based learning on stance detection in Section \ref{section:promptbasedlearningonsd}.
\subsection{Stance Detection} \label{section:stancedetection}
Stance detection aims to identify the stance toward a specific target. Previous works mainly focus on identifying stance in debates \cite{thomas2006get,somasundaran2009recognizing,murakami2010support} or forums \cite{sridhar2014collective}. The stance detection tested in earlier work includes rule-based algorithms and feature-based machine learning approaches \cite{tutek2016takelab,bohler2016idi}. Recently, there has been a growing interest in performing stance detection on social media. Mohammad \textit{et al.} \cite{mohammad2016semeval} released the SemEval-2016 task 6 dataset, one widely known benchmark dataset for stance detection derived from Twitter. Various models based on deep neural networks (such as RNNs with their modified versions and CNNs) have been proposed for stance detection \cite{rajendran2018stance, hercig2017detecting}. Recently, it has been a common practice to combine deep learning along with attention methods to learn target-aware representations. Du \textit{et al.} \cite{du2017stance} incorporated target-specific information into stance detection by following an attention mechanism. Dey \textit{et al.} \cite{dey2018topical} proposed a two-phase LSTM based model with attention and Siddiqua \textit{et al.} \cite{siddiqua2019tweet} adopted two LSTM variants where each module was coupled with an attention mechanism.

However, due to the lack of context semantic information between text and target, it is not enough to learn latent representation and relationship between target and text only through the above method. Some studies supplement the semantic information by introducing external knowledge bases. Du \textit{et al.} \cite{du2020commonsense} designed a commonsense knowledge enhanced memory network, which jointly represented textual and commonsense knowledge representation of given target and text. Hanawa \textit{et al.} \cite{hanawa2019stance} presented a method of extracting related concepts from Wikipedia articles and incorporated extracted knowledge into stance detection. Zhang \textit{et al.} \cite{zhang2020enhancing} proposed a semantic-emotion knowledge transferring model for cross-target stance detection, which used external knowledge as a bridge to enable knowledge transfer across different targets. Liu \textit{et al.} \cite{ liu2021enhancing} introduced a commonsense knowledge enhanced model to exploit both the structural-level and semantic-level information of the relational knowledge. Besides, Zhang \textit{et al.}  \cite{zhang2021knowledge} leveraged multiple external knowledge bases as bridges to explicitly link potentially opinioned terms in texts to targets of interest. 

Although these methods alleviate the lack of context information, there is no significant increase in performance. Part of the reason is that introducing existing knowledge bases in stance detection may bring noise. At the same time, poor entity linking and fewer related entries might lead to limited number of words per text with external information. Some studies supplement contextual semantic information by fine-tuning pre-trained language models (\textit{e.g.}, GPT \cite{radford2018improving}, BERT \cite{devlin2019bert}, RoBERTa \cite{liu2019roberta}, ALBERT \cite{lan2019albert}), yielding state-of-the-art performances. Popat \textit{et al.} \cite{popat2019stancy} leveraged fine-tuning BERT representations and augmented them with consistency constraints. S{\'a}enz \textit{et al.} \cite{ saenz2021interpreting} addressed stance detection by proposing a BERT-based classification model and an attention-based mechanism to identify the influential words for stance classification. Additionally, Sun \textit{et al.} \cite{ sun2021stance} proposed a knowledge-enhanced BERT model in which triples in knowledge graphs were used as domain knowledge injected into the text. 

\subsection{Prompt-based Learning} \label{section:prompt-basedlearning}
Most existing PLMs are pre-trained with the standard ``pre-training and fine-tuning'' paradigm. However, the gap between the pre-training stage and the downstream task can be significant, which hinders the exploitation of the knowledge implicit in the PLM. To bridge the gap, prompt-based learning has been introduced. In this paradigm, instead of adapting PLMs to downstream tasks via objective engineering, downstream tasks are reformulated to look more like those solved during the original LM training with the help of a textual prompt \cite{liu2021pre}. Notably, prompt-based PLM has shown significant potential among different tasks, such as classification-based tasks, information extraction, question answering, and text generation. In this paper, we focus on classification-based tasks, so we introduce some works which utilize prompt-based learning to solve classification. Yin \textit{et al.} \cite{yin2019benchmarking} used a prompt such as “the topic of this document is [Z].”, which was then fed into mask PLMs for slot filling. Schick \textit{et al.} \cite{schick2021exploiting} introduced Pattern-Exploiting Training (PET), a semi-supervised training procedure that reformulated input examples as cloze-style phrases to help language models understand a given task. Seoh \textit{et al.} \cite{seoh2021open} resolved aspect target sentiment classification by treating the review as a premise and the prompt sentence with the sentimental next word as a hypothesis and predicting whether the review entailed the prompt.

In this paper, we propose a target-aware verbalizer and multi-prompts distillation inspired by the format of cloze questions using PET.

\subsection{Prompt-based Learning on Stance Detection} \label{section:promptbasedlearningonsd}

There have been two studies on stance detection, which focus on the multilingual setting. Schick \textit{et al.} \cite{schick2021exploiting} evaluated the PET method by X-stance dataset, which is a multilingual binary stance detection dataset with German, French and Italian examples. Hardalov \textit{et al.} \cite{hardalov2021few} added a label encoder to PET and  utilized sentiment-based generation of stance data for pre-training to resolve cross-lingual stance detection. In this work, we focus on English stance detection.

\begin{figure*}[h]
  \centering
  \includegraphics[width=0.75\linewidth]{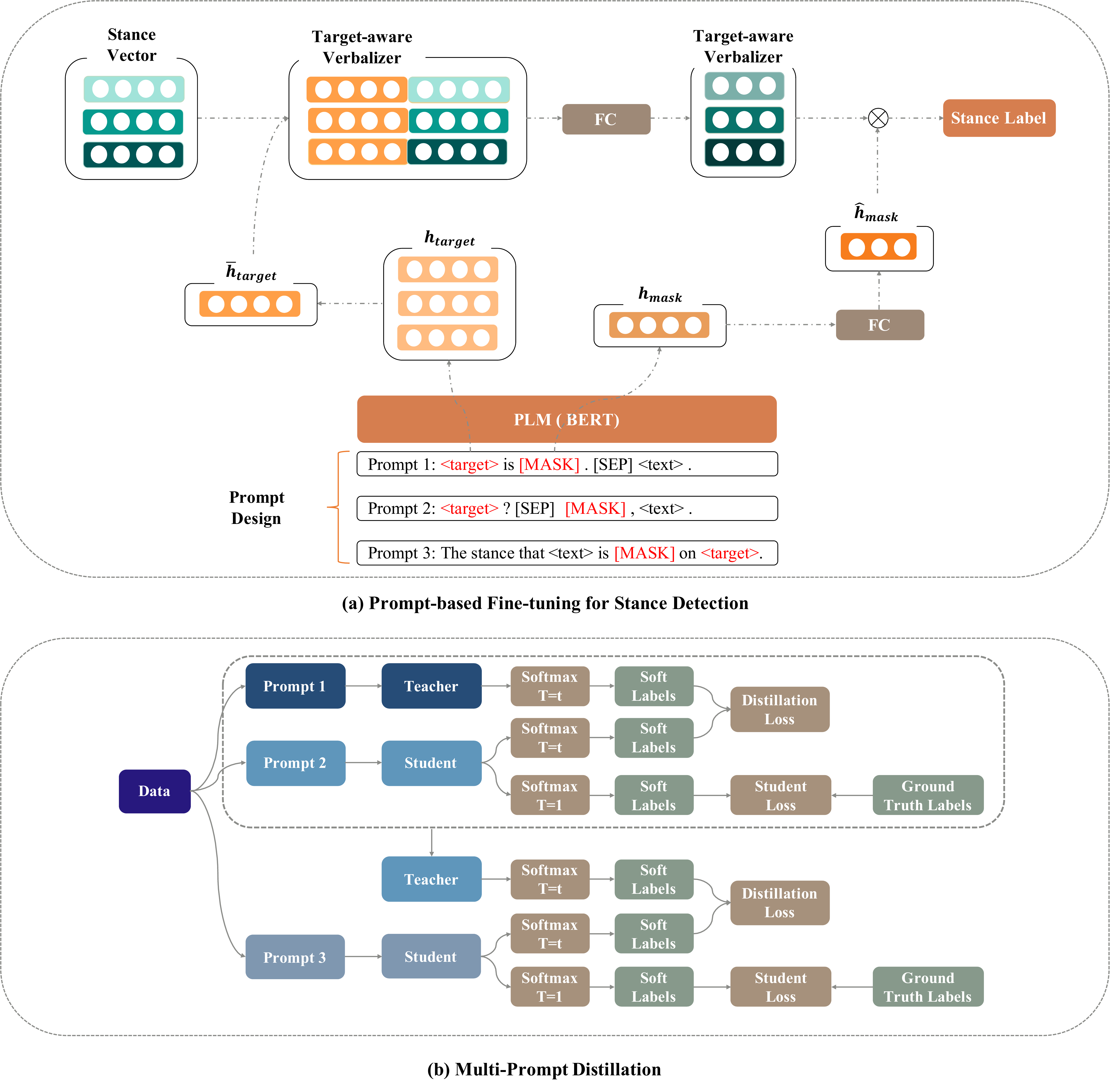}
  \caption{Prompt-base fine-tuning and multiple prompt distillation.}
    \label{fig:model}
\end{figure*}

\section{Model}
In this section, we present our proposed  Target-Aware Prompt Distillation (TAPD) for stance detection in detail. We first provide the formulation of the stance detection task in Section \ref{section:problemstatement}, and then describe the detailed implementation of our proposed framework. As demonstrated in Figure \ref{fig:model}, the architecture of our proposed TAPD framework consists of two main components: 1) prompt-based fine-tuning for stance detection, in which we present our customized verbalizer to capture the correlation between the stance and the target for stance detection (described in Section \ref{section:prompt-basedstancedetection}); 2) multi-prompt distillation which learns information from diverse prompts (described in Section \ref{section:promptdistillation}). Finally, we introduce the training and inference of our model \ref{section:trainingandinference}.

\subsection{Problem Formulation} \label{section:problemstatement}
Given a sentence composed of $n$ words \begin{math}
  s = \{w_1,w_2,...,w_n\} 
\end{math}, and a target composed of $m$ words \begin{math} t = \{c_1,c_2,..,c_m\}
\end{math}, the task of stance detection aims to identify the stance \begin{math} y
\end{math} that $s$ expresses towards $t$. The stance label set \begin{math}
\mathcal{Y}
\end{math} consists of \textit{Favor}, \textit{Against}, and \textit{None}.

It is noteworthy that there exist two training settings for stance detection. One is per-target training, \textit{i.e.}, training a separate model for each target and evaluating the model on texts related to that target, while multi-target training trains one model for all the targets. Per-target training neglects the role of the target when making predictions and may exacerbate the label scarcity issue under the few-shot setting. In this paper, we adopt the multi-target training setting.

\subsection{Prompt-based Fine-tuning for Stance Detection} \label{section:prompt-basedstancedetection}
In this section, we describe the details of our proposed prompt-based fine-tuning for stance detection. It consists of prompt design (described in Section \ref{section:promptsdesin}), target-aware verbalizer (described in Section \ref{section:targetadaptive}), and stance classification (described in Section \ref{section:stancerepresentation}).

\subsubsection{Prompt Design} \label{section:promptsdesin}

The key to prompt-based fine-tuning for classification tasks lies in reformulating it as an appropriate prompt, which is also applicable to stance detection. Previous works have shown that the performance of different prompts varies significantly \cite{le2021many}
and the situation becomes even worse for stance detection. The expressions and topics are quite different among various targets, making it impractical to construct a universal prompt for all the targets. Therefore, we construct three prompts from different perspectives, \textit{i.e.}, stance detection as sentiment analysis \cite{seoh2021open}, stance detection as natural language inference \cite{schick2021exploiting}, and stance detection task itself \cite{hardalov2021few}. To adapt these prompts to stance detection, we make some slight adjustments. For each input pair \begin{math}
  x = (target,text)
\end{math}, we define the following three patterns:
\begin{align*}
& P_1(target,text) \ = \ \left \langle target \right \rangle\ {is\ [MASK].}\ {[SEP]} \ \left \langle text \right \rangle. \\
& P_2(target,text) \ = \ \left \langle target \right \rangle\ {?\ [SEP]\ [MASK] \ ,\ } \left \langle text \right \rangle.  \\ 
& P_3(target,text) \ = \ {The\ stance\ that\ } \left \langle text \right \rangle   {\ is\ [MASK] \ on\ } \left \langle target \right \rangle . 
\end{align*}
We take BERT \cite{devlin2019bert} as our PLM and the $[MASK]$ and $[SEP]$ tokens are directly taken from the BERT vocabulary. Our designed prompts can be easily adapted to the pre-training tasks of other PLMs.

\subsubsection{Target-aware Verbalizer}
\label{section:targetadaptive}

Prompt-based fine-tuning usually define a verbalizer, an injective function  \begin{math}
f:\mathcal{Y} \to V
\end{math}, to map each label to a single token from the PLM's vocabulary \cite{schick2021exploiting}, \textit{e.g.}, \begin{math} ``Favor" \to ``Yes", \ ``Against" \to ``No", \ ``None" \to ``Maybe" \end{math} for our constructed prompt $P_2$. The design of the verbalizer has proved to be crucial for prompt-based methods \cite{gao2020making}

and simply mapping each label to a pre-defined concrete word might not work. Schick \textit{et al.} \cite{schick2020s} tried to map each label into a phrase to better capture the semantic meaning of the labels, \textit{e.g.}, ``in favor of'' instead of ``favor''. Though sounding promising, predicting consecutive $[mask]$ tokens brings a new challenge. 
Gao \textit{et al.} \cite{gao2020making} constructed a pruned set of the top $k$ vocabulary words that better generate the label via the PLM and automatically searched the verbalized word for each label. However, brute-forcing label searching is computationally intensive and time-consuming. Moreover, due to the diverse expressions among various targets, we argue that only a single token or phrase may be insufficient to capture the stance information. To address this issue, we propose to map the labels to continuous vectors instead of concrete words or phrases, which are called stance vectors and are trainable during optimization. We construct three stance vectors, namely \begin{math} V_{favor},\ V_{against},\end{math} and \begin{math} V_{none} \end{math} for \textit{Favor}, \textit{Against}, and \textit{None}, respectively. The dimension of stance vectors is set to be the same as the size of token embeddings in the PLM.

To capture the strong interplay between the target and the stance, we propose to learn target-aware stance labels for each target. We first calculate the representation of the target. In our model, the prompt is first fed into the pre-trained language model. The PLM computes the latent representations for all the tokens in the sequence, including the $[mask]$ token and the $[target]$ tokens, written as \begin{math} h_{target} \in \mathbb{R}^{d_h} \end{math} and \begin{math} \ h_{mask} \in \mathbb{R}^{d_h} \end{math} where $d_h$ is the dimension of the hidden states in the pre-trained language model.  As the target may consist of more than one single token, we apply average pooling to the tokens in the target to obtain a more compact target representation $h_{target}$:
\begin{equation}
    h_{target}=\frac{1}{m}\sum_{i=1}^{m}h_{i},  
\end{equation}
where \begin{math}h_{i} \in  \mathbb{R} ^{d_h} \end{math} is the $d_h$-dimensional vector of the $i$-th token in the target sequence and $m$ is the total number of target tokens. 

After obtaining the target representation, we directly concatenate the target representation and the stance vector as the target-aware stance vector: 
\begin{equation}
\begin{aligned}
     {VT}_{favor} &= h_{target} \oplus V_{favor}, \\
     {VT}_{none} &= h_{target} \oplus V_{none}, \\
     {VT}_{against} &= h_{target} \oplus V_{against},
\end{aligned}
\end{equation}
where $\oplus$ is the vector concatenation operation. The dimension of $VT$ is $2d_h$.

\subsubsection{Stance Classification} \label{section:stancerepresentation}
We classify the stance expressed in the text by measuring the semantic similarity between the $[mask]$ token and the target-aware stance vectors. We expect that the $[mask]$ token has collected sufficient contextual information of the target and that the prompts have guided the PLM to associate the accurate stance with this token. As their dimensions differ, we use two linear transformations to map $h_{mask}$ and $VT$ into $\hat{h}_{mask}$ and $\hat{VT}$, both of which have a dimension of ${d_m}$. To infer the label for each text, we take the dot product between $h_{mask}$ and $VT$ and 
apply the softmax to obtain the probability of each stance:
\begin{equation}
\label{eq:classification}
     p(y_i|s_{prompt}) = \frac{ exp(\hat{h}_{mask} \cdot \hat{VT}_{i})}{ \sum\limits_{{\hat{VT}}_{t}\in {VT}} exp(\hat{h}_{mask} \cdot \hat{VT}_{t})}, 
\end{equation}
where \begin{math} VT=\{ {\hat{VT}}_{favor}, {\hat{VT}}_{none}, {\hat{VT}}_{against} \} \end{math} denotes the set of target-aware stance vectors.

\subsection{Multi-Prompt Distillation}  \label{section:promptdistillation}
As mentioned before, we design three prompts from different perspectives to cover diverse targets. However, how to merge the results of these three prompts needs careful design. Existing methods simply average the output label propensity score \cite{schick2021exploiting}, neglecting the intrinsic knowledge learned by each prompt. In this paper, we propose to utilize knowledge distillation mechanism \cite{hinton2015distilling} to distill the knowledge inhabited in prompt-based fine-tuned PLMs. 

As shown in Figure \ref{fig:model}(b), our distillation framework follows a sequential manner. At each time step in distillation, the current model (called student model) is trained with two teachers: ground-truth labels and soft labels generated by the teacher model which is the fine-tuned student model at the previous time step. At the first time step, the student model is trained with only the ground-truth labels. The student model trained at the last time step is taken as our final classification model. Student models at different time steps are fine-tuned with different prompts and the order of student models shows no significant impact during our experiments.

At each time step, the student model needs to align its output with two teachers: ground-truth labels and soft labels generated by the teacher model. The student model infers the stance labels as described in Section \ref{section:prompt-basedstancedetection} and optimizes the distance between predicted labels and ground-truth labels. Meanwhile, both the teacher model and the student model are applied to make predictions under temperature $T$ and the distance between their predictions is optimized. The temperature $T$ helps reweight the importance of different classes and is a common practice in model distillation \cite{hinton2015distilling}. The adjusted prediction probability for both the teacher model and the student model is calculated as follows:

\begin{equation}
    p(y_i|s_{prompt})_{soft} = \frac{ exp((\hat{h}_{mask} \cdot \hat{VT}_{i}) / T )}{ \sum\limits_{{\hat{VT}}_{t}\in {VT}} exp((\hat{h}_{mask} \cdot \hat{VT}_{t})/T)}.
\end{equation}

After optimization, the student model will serve as the teacher model to guide the distillation for the next time step.

\subsection{Training and Inference} \label{section:trainingandinference}
The overall framework is trained in a sequential manner for multi-prompt distillation. At each time step, the overall training objective consists of two parts: the stance classification loss \begin{math} \mathcal{L}_C  \end{math} and the distillation loss \begin{math} \mathcal{L}_D \end{math}. The stance classification loss aims at measuring the distance between the predicted labels and ground-truth labels and is defined as a standard cross entropy loss:
\begin{equation}
    \mathcal{L}_C = \sum_{x \in D} CE(y_x^{pred},y_x^{true}),
\end{equation}
where $CE$ denotes the cross-entropy loss function, $y_x^{pred}$ and $y_x^{true}$ are the predicted label and ground-truth label for the sample $x$ in training set $D$. The distillation loss measures the distance between the soft labels generated by the teacher model and the student model and is defined in a similar way.

The overall objective at each time step is defined as the sum of \begin{math} \mathcal{L}_C  \end{math} and \begin{math} \mathcal{L}_D  \end{math}. It is calculated as follows:
\begin{equation}
    \mathcal{L} = \lambda \cdot \mathcal{L}_C + (1-\lambda)\cdot T^2 \cdot \mathcal{L}_D, 
\end{equation}
where $\lambda$ is a hyper-parameter to balance the importance of these two losses and $T$ is the temperature of distillation. As suggested in previous works \cite{hinton2015distilling}, we multiply the distillation loss $\mathcal{L}_D$ by $T^2$ since the magnitudes of the gradients produced by the distillation loss is scaled by the factor $1/T^2$ during optimization.

When making inference, we take the student model trained at the last time step as our prediction model. For an input text, we inject it into the prompt used in the prediction model and feed the filled prompt into the PLM. We calculate the prediction probability according to Equation (\ref{eq:classification}) and pick the label with the highest prediction probability.

\section{Experiments}

In this section, we conduct extensive experiments on two widely adopted datasets for stance detection to validate the effectiveness of our proposed model. We first introduce our experimental setup and baseline methods in Section \ref{experimentalsetup} and Section \ref{baseline}. The experimental results are presented in Section \ref{results} with in-depth analysis and interpretation. Finally, we thoroughly analyze the details of our model design in Section \ref{analysis}.

\subsection{Experimental Setup} \label{experimentalsetup}

In this section, we provide a comprehensive description of our experimental setup, including the stance detection datasets adopted for evaluation (described in Section \ref{datasets}), the evaluation metrics (described in Section \ref{evaluationmetrics}) and the implementation details of our model (described in Section \ref{implementationdetails}).

\subsubsection{Datasets} \label{datasets}
We test our approach on SemEval-2016 Task 6 Sub-task A dataset and UKP dataset, both of which are widely adopted in previous stance detection research. The examples of the SemEval-2016 dataset and the UKP dataset are shown in Table~\ref{tab:dataexample}.

\begin{table}
  \caption{Examples from datasets.}
  \label{tab:dataexample}
  \resizebox{\linewidth}{!}{
  \begin{tabular}{lp{1.5cm}p{5cm}p{1cm}}
    \toprule
    \textbf{Dateset} & \textbf{Target} & \textbf{Text} & \textbf{Stance} \\
    \midrule
    SemEval-2016 & Atheism & God of the gaps is not evidence \#next \#SemST & Favor \\
    \midrule
    SemEval-2016 & Feminist Movement & I'm not sure my schadenfreudes can stand this much tickling. @MT8\_9 @prettysing \#SemST & None \\
    \midrule
    SemEval-2016 & Hillary Clinton & @HillaryClinton The deceit hand? Thats a hand you know well. Right? \#SemST & Against \\
    \midrule
    UKP & death penalty &and look , it is about justice & Favor \\
    \midrule
    UKP &cloning & what is stem cell research ? & None\\
    \midrule
    UKP & Abortion & If the fetus developed the sense of pain, than it is a shame. & Against \\
  \bottomrule
\end{tabular}}
\end{table}

The SemEval-2016 dataset \cite{mohammad2016semeval} consists of 4,163 English tweets crawled from Twitter. Each tweet is assigned with a target and a manually annotated stance label (favor, neutral, or against) towards the given target. There are five different targets in this dataset: ``Atheism (AT)'', ``Climate Change is a real Concern (CC)'', ``Feminist Movement (FM)'', ``Hillary Clinton (HC)'' and ``Legalization of Abortion (LA)''. The detailed data distribution of the SemEval-2016 dataset is shown in Table~\ref{tab:semevaldataset}. As in previous works, we adopt the official train/test split. Since the task did not provide an official validation set, we split the train set in a 5:1 ratio into train and validation sets.

\begin{table}
  \caption{Data distribution of the SemEval-2016 dataset.}
  \label{tab:semevaldataset}
  \resizebox{\linewidth}{!}{
  \begin{tabular}{ccccccccc}
    \toprule
    \textbf{Target} & \textbf{\#Train} & \%Favor & \%Against & \%None & \textbf{\#Test} & \%Favor  & \%Against & \%None \\
    \midrule
    \textbf{AT} & 513 & 17.93 & 59.26 & 22.81 & 220 & 14.54 & 72.73 & 12.73 \\
    \textbf{CC} & 395 & 53.67 & 3.80 & 42.53 & 169 & 72.78 & 6.51 & 20.71\\
    \textbf{FM} & 664 & 31.63 & 49.40 & 18.97 & 285 & 20.35 & 64.21 & 15.44\\
    \textbf{HC} & 689 & 17.13 & 57.04 & 25.83 & 295 & 15.25 & 58.31 & 26.44\\
    \textbf{LA} & 653 & 18.53 & 54.36 & 27.11 & 280 & 16.43 & 67.50 & 16.07\\
    \midrule
    \textbf{Total} & 2,914 & 25.84 & 47.87 & 26.29 & 1,249 & 24.34 & 57.25 & 18.24 \\
  \bottomrule
\end{tabular}}
\end{table}

The UKP dataset \cite{stab2018cross} consists of 25,492 argument sentences from 400 internet texts (from essays to news text) on eight different topics, including ``Abortion (AB)'', ``Cloning (CL)'', ``Death Penalty (DP)'', ``Gun Control (GC)'', ``Marijuana Legalization (ML)'', ``Minimum Wage (MW)'', ``Nuclear Energy (NE)'' and ``School Uniforms (SU)''. The dataset is annotated to detect whether an argument is in support of, neutral, or opposed to a given target. We adopt the train, validation, and test splits provided by the authors. The detailed data distribution of the UKP dataset is shown in Table~\ref{tab:ukp-dataset}.

\begin{table*}
  \caption{Data distribution of the UKP dataset.}
  \label{tab:ukp-dataset}
  \resizebox{\linewidth}{!}{
  \begin{tabular}{ccccccccccccc}
    \toprule
    \textbf{Target} & \textbf{\#Train} & \%Favor & \%Against & \%None & \textbf{\#Val} & \%Favor & \%Against & \%None & \textbf{\#Test} & \%Favor & \%Against & \%None \\
    \midrule
    \textbf{AB} & 2,827  & 17.33 & 20.91 & 61.76 & 315 & 17.14 & 20.95 & 61.90 & 787 & 17.28 & 20.97 & 61.75  \\
    \textbf{CL} & 2,187  & 23.23 & 27.62 & 49.15 & 243 & 23.05 & 27.57 & 49.38 & 609 & 23.32 & 27.59 & 49.10 \\
    \textbf{DP} & 2,627  & 12.03 & 30.03 & 57.94 & 293   & 12.97 & 30.72 & 56.31 & 731   & 14.09 & 31.74 & 54.17 \\
    \textbf{GC} & 2,404  & 23.54 & 19.93 & 56.53 & 268   & 23.51 & 19.78 & 56.72 & 669   & 23.62 & 19.88 & 56.50 \\
    \textbf{ML} & 1,780  & 23.71 & 25.28 & 51.01 & 198   & 23.74 & 25.25 & 51.01 & 497   & 23.74 & 25.35 & 50.91 \\
    \textbf{MW} & 1,778  & 23.28 & 22.27 & 54.44 & 198   & 23.23 & 22.22 & 54.55 & 497   & 23.34 & 22.33 & 54.33 \\
    \textbf{NE} & 2,573  & 16.95 & 23.82 & 59.23 & 286   & 16.78 & 23.78 & 59.44 & 717   & 17.02 & 23.85 & 59.14 \\
    \textbf{SU} & 2,165  & 18.11 & 24.25 & 57.64 & 241   & 18.26 & 24.07 & 57.68 & 602   & 18.11 & 24.25 & 57.64  \\
    \midrule
    \textbf{Total} & 18,341 & 19.32 & 24.25 & 56.43 & 2,042  & 19.39 & 24.29 & 56.32 & 5,109  & 19.65 & 24.51 & 55.84 \\
  \bottomrule
\end{tabular}}
\end{table*}

\subsubsection{Evaluation Metrics} \label{evaluationmetrics}
Following previous stance detection works, we adopt macro-average of F1 score ( denoted as $MacF_{avg}$) and micro-average of F1 score (denoted as $MicF_{avg}$) to evaluate the performance of the proposed model, as in Mohanmmad \textit{et al.} \cite{mohammad2016semeval} and Li \textit{et al.} \cite{li2019multi}. The evaluation focuses on the \textit{Favor} class and the \textit{Against} class. Firstly, the F1 score of the \textit{Favor} class and the \textit{Against} class is calculated as follows:
\begin{equation}
    F_{favor} = \frac{2 \times P_{favor} \times R_{favor}}{P_{favor}+R_{favor}}, 
\end{equation}
\begin{equation}
    F_{against} = \frac{2 \times P_{against} \times R_{against}}{P_{against}+R_{against}}, 
\end{equation}
where $P_{\cdot}$ and $R_{\cdot}$ are the precision and recall of the corresponding class. Then the F1 average is calculated as:
\begin{equation}
    F_{avg} = \frac{F_{favor}+F_{against}}{2}.
\end{equation}
Note that the \textit{None} class is not discarded during training. However, the \textit{None} class is not considered in the evaluation because we are only interested in the \textit{Favor} class and the \textit{Against} class in this task.

To calculate $MacF_{avg}$, we first generate the $F_{avg}$ on each target and take the average of $F_{avg}$ over all the targets. To get $MicF_{avg}$, we first calculate $F_{favor}$ and $F_{against}$ across all the targets and take their average as the final measure.

\subsubsection{Implementation Details} \label{implementationdetails}
We implement our approach using PyTorch 1.4.0 and train our model on a Tesla V100 GPU. We use the BERT base model (bert-base-uncased) as our pre-trained language model. It consists of 12 transformer layers, each of which adopts a hidden state size of 768 and 12 attention headers. We fine-tune our model with Adam optimizer. The dropout rate is set to 0.5 for all the parameters. The learning rate is chosen from $\{ 5\times10^{-5},1\times10^{-5},5\times10^{-6},1\times10^{-6}\}$ and set to $1\times10^{-5}$. The batch size for training is set to 32. The maximum sequence length is set to 128 in the SemEval-2016 dataset and 200 in the UKP dataset. $d_m$ is set to 384. 
The temperature is set to 2 and $\lambda$ is set to 0.8.

\subsection{Baseline Methods} \label{baseline}
Models from three typical categories are chosen as our baselines: attention-based sequential models, BERT-based fine-tuning models and models with external knowledge. The detailed baselines are listed as follows:
\begin{itemize}
\item {\verb|TAN| (\cite{du2017stance})}: An LSTM-attention based model that leverage the attention module to incorporate target-specific information into stance detection. 
\item {\verb|JOINT| (\cite{sun2019stance})}: A joint model that exploited sentiment information to improve stance detection.
\item {\verb|PNEM| (\cite{siddiqua2019tweet})}: An ensemble model that adopted two LSTM-attention based models to learn long-term dependencies and a multi-kernel convolution to extract the higher-level tweet representation.
\item {\verb|AT-JSS-Lex| (\cite{li2019multi})}: A multi-task framework that used a sentiment lexicon and constructed a stance lexicon to guide target-specific attention mechanism. Besides, it took sentiment classification as an auxiliary task.
\item {\verb|CKEMN| (\cite{du2020commonsense})}: A commonsense knowledge enhanced memory network for stance detection using LSTM as embedding.
\item {\verb|RelNet| (\cite{zhang2021knowledge})}: A multiple knowledge enhanced framework for stance detection using BERT.
\item {\verb|BERT|$_{SEP}$ (\cite{devlin2019bert})}: A pre-trained language model that predicted the stance by appending a linear classification layer to the hidden representation of [\textit{CLS}] token. We fine-tune the BERT on the stance detection.
\item {\verb|BERT|$_{MEAN}$ (\cite{ma2019universal})}: A pre-trained language model that predicted the stance by appending a linear classification layer to the mean hidden representation over all tokens. We fine-tune the BERT on the stance detection.
\item {\verb|BERT|$_{TAN}$}: A variant of TAN, in which we replace the LSTM features with BERT generated word embeddings.
\item {\verb|Stancy| (\cite{popat2019stancy})}: A BERT based model that leveraged BERT representations learned over massive external corpora and utilized consistency constraint to model target and text jointly.
\item {\verb|S-MDMT| (\cite{wang2021solving})}: A BERT based model that applied the target adversarial learning to capture stance-related features shared by all targets and combined target descriptors for learning stance-informative features correlating to specific targets.
\end{itemize}
In the SemEval-2016 dataset, we report the results of all the baseline methods mentioned above. Note that the results for Stancy\footnote{\url{https://github.com/kashpop/stancy}.}, BERT$_{SEP}$, BERT$_{MEAN}$ and BERT$_{TAN}$ are collected through our experiments and the results of other models are directly taken from the original paper.
For the UKP dataset, the results of some baselines are unavailable and the baselines cannot be reproduced due to additional lexicons. Therefore, we report the results of four baselines that perform well in the SemEval-2016 dataset. The four baselines are Stancy, BERT$_{SEP}$, BERT$_{MEAN}$, and BERT$_{TAN}$. All the hyper-parameters of baselines are selected by grid search on the validation set.

\subsection{Results} \label{results}
In this section, we present the experimental results in detail. Our experiments try to investigate the following research questions:
\begin{itemize}
\item \textbf{RQ1}: Can TAPD better capture the contextual information of the targets?
\item \textbf{RQ2}: Can TAPD benefit few-shot stance detection?
\end{itemize}
To achieve this goal, we conduct stance detection in both full data and few-shot scenarios respectively.

\begin{table}
  \centering
  \caption{Full-data stance detection performance on the SemEval-2016 dataset.}
    \label{tab:semevalperformance}
     \resizebox{\linewidth}{!}{
    \begin{tabular}{cccccccc}
    \toprule
    Model & AT & CC & FM & HC & LA & {$MacF_{avg}$} & {$MicF_{avg}$} \\
    \midrule
    TAN \cite{du2017stance})   & 59.33 & 53.59 & 55.77 & 65.38 & 63.72 & 59.56 & 68.79 \\
    JOINT \cite{sun2019stance} & 66.78 & 50.6  & 59.35 & 62.47 & 61.58 & 60.16 & 69.22 \\
    PNEM \cite{siddiqua2019tweet} & 67.73 & 44.27 & \textbf{66.76} & 60.28 & 64.23 & 60.65 & 72.11 \\
    AT-JSS-Lex \cite{li2019multi} & 69.22 & 59.18 & 61.49 & 68.33 & \textbf{68.41} & 65.33 & 72.33 \\
    \midrule
    CKEMN \cite{du2020commonsense} & 62.69 & 53.52 & 61.25 & 64.19 & 64.19 & 61.17 &69.74 \\ 
    RelNet \cite{zhang2021knowledge} & 70.55 & 57.20 & 61.25 & 62.33 & 63.65 & 63.06 & 71.06 \\
    \midrule
    BERT$_{SEP}$ \cite{devlin2019bert} & 68.67  & 44.14  & 61.66  & 62.34  & 58.60  & 59.09 & 69.51 \\
    BERT$_{MEAN}$ \cite{ma2019universal} & 69.44 & 52.47 & 59.22  & 64.61  & 66.30  & 62.41  & 70.92  \\
    BERT$_{TAN}$  & 65.51  & 58.55  & 58.30  &  64.31  & 63.58  & 62.05  & 70.28 \\
    Stancy \cite{popat2019stancy} & 69.85 & 53.47 & 61.67  & 64.70 & 63.42 & 62.62 & 71.77 \\
    S-MDMT \cite{wang2021solving} & 69.50  & 52.49 & 63.78 & 67.20  & 67.19 & 64.03 & 72.70 \\
    \midrule
     TAPD  & \textbf{73.87}  & \textbf{59.32} & 63.93  & \textbf{70.01} & 67.23 &  \textbf{66.87} & \textbf{74.80}  \\
    \bottomrule
    \end{tabular}}
\end{table}

\begin{table}
  \centering
  \caption{Full-data stance detection performance on the UKP dataset.}
    \label{tab:ukpperformance}
    \resizebox{\linewidth}{!}{
    \begin{tabular}{ccccccccccc}
    \toprule
    {Model} & AB & CL & DP & GC & ML & MW & NE & SU & {$MacF_{avg}$} & {$MicF_{avg}$} \\
    \midrule
    Stancy & 51.76  & 67.52  & 56.71  & 50.24  & 65.35  & 66.15  & 58.32  & 62.91  & 59.87  & 60.06 \\
    BERT TAN &  40.57  & 69.36  & 51.63  & 50.98  & 63.47  & 64.10  & 56.70  & 64.08  & 57.61  & 57.94  \\
    BERT SEP & 49.09  & 66.90  & 52.42  & 51.62  & 66.31  & 64.91  & 58.54  & 63.42  & 59.15  & 59.73  \\
    BERT MEAN &  53.76  & 68.55  & 54.16  & 51.12  & 65.18  & 66.40  & 57.76  & 63.27  & 60.03  & 60.19  \\
    \midrule
    TAPD   &  \textbf{54.87}  & \textbf{71.86}  & \textbf{57.16}  & \textbf{51.73}  & \textbf{67.01}  & \textbf{69.95}  & \textbf{59.03}  & \textbf{64.10}  & \textbf{61.96}  & \textbf{62.15} \\
    \bottomrule
    \end{tabular}}
\end{table}

\subsubsection{Full Data Learning}
Table \ref{tab:semevalperformance} and Table \ref{tab:ukpperformance} show the results of different methods on the SemEval2016 dataset and the UKP dataset, respectively. We can observe that our model outperforms all the baseline models in terms of overall $MacF_{avg}$ and $MicF_{avg}$ on both datasets. It also achieves the best or comparable performance on each target. Moreover, BERT-based models generally achieve better performance, demonstrating that PLMs can provide more effective embeddings. However, there exists a gap between the training and inference of the traditional fine-tuning paradigm. In contrast, prompt-based fine-tuning can bridge such a gap by reformulating the downstream task as a language modeling problem, thus leveraging the implicit knowledge learned during pre-training. Therefore, our proposed TAPD significantly outperforms BERT-based fine-tuning methods.

\subsubsection{Few-shot Learning}
We evaluate the performance of few-shot learning with a various number of training examples on both datasets. Within each label of each target, we randomly collect 2, 5, 10, 20, 30 examples respectively, resulting in 30, 75, 150, 300, and 450 training samples on the SemEval-2016 dataset and 48, 120, 240, 480, and 720 training samples on the UKP dataset. We repeat the experiment five times and report the average results. We select four baselines, \textit{i.e.}, Stancy, BERT$_{SEP}$, BERT$_{MEAN}$ and BERT$_{TAN}$. Table \ref{tab:fewshotperformance} and Table \ref{tab:fewshotukp} show the results of few-shot learning on both datasets. We can see that TAPD achieves better performance than all the baselines in terms of overall $MacF_{avg}$ and $MicF_{avg}$. The fewer training samples, the larger performance improvement for TAPD, especially on the UKP dataset. This phenomenon demonstrates the capability of TAPD to handle the few-shot learning scenario where only a few training samples are available.

\begin{table}
  \centering
  \caption{Few-shot stance detection performance on the SemEval-2016 dataset.}
    \label{tab:fewshotperformance}
    \resizebox{\linewidth}{!}{
    \begin{tabular}{cccccccc}
    \toprule
    {Model} & AT & CC & FM & HC & LA & {$MacF_{avg}$} & {$MicF_{avg}$} \\
    \midrule
   train data (30) \\
    Stancy &  45.62  & 10.59  & 37.83  & 30.46  & \textbf{41.30}  & 33.16($\pm$4.68)  & 39.40($\pm$5.38) \\
    BERT$_{SEP}$ & 37.74 & 6.99  & 38.38 & \textbf{36.66}  & 37.53  & 31.46($\pm$2.76)  & 34.98($\pm$1.19)  \\
    BERT$_{MEAN}$ & 32.17& 16.11  & 40.68  & 34.90  & 37.72  & 32.32($\pm$3.87)  & 36.38($\pm$3.84) \\
    BERT$_{TAN}$ &  42.38  & 32.39  & 36.22  & 30.77  & 36.48  & 35.65($\pm$2.80)  & 39.88($\pm$3.92)  \\
    TAPD  & \textbf{46.51}  & \textbf{34.84} & \textbf{41.05}  & 35.05  & 40.64  & \textbf{39.62}($\pm$2.18)  & \textbf{45.23}($\pm$4.83) \\
    \midrule
    train data (75) \\
    Stancy & 47.07  & 34.36  & 40.21  & 33.02  & 42.36  & 39.41($\pm$7.48)  & 41.67($\pm$8.61) \\
    BERT$_{SEP}$ & 37.91  & 36.77  & \textbf{44.53}  & 40.17  & 35.80  & 39.04($\pm$2.79)  & 43.56($\pm$1.50) \\
    BERT$_{MEAN}$ &  46.70  & 33.71  & 42.78  & 39.84  & 44.22  & 41.45($\pm$2.64)  & 46.13($\pm$2.61)  \\
    BERT$_{TAN}$ & 44.89  & 36.73  & 39.42  & 39.76  & 41.85  & 40.53($\pm$4.20)  & 44.42($\pm$2.81) \\
    TAPD & \textbf{48.04}  & \textbf{36.83}  & 43.60  & \textbf{48.07}  & \textbf{44.82}  & \textbf{44.27}($\pm$2.61)  & \textbf{49.07}($\pm$3.93) \\
    \midrule
    train data (150) \\
    Stancy &  \textbf{52.37} & 44.18  & 40.51  & 40.26  & 42.79  & 44.02($\pm$2.56)  & 48.51($\pm$3.23) \\
    BERT$_{SEP}$ & 50.15  & 44.36  & 42.23  & 38.76  & 45.79  & 44.26($\pm$2.57)  & 50.17($\pm$2.29)  \\
    BERT$_{MEAN}$ & 48.70  & 45.97  & \textbf{48.05}  & 42.50  & 40.01  & 45.04($\pm$4.48)  & 50.08($\pm$3.68) \\
    BERT$_{TAN}$ & 48.55  & 44.82  & 44.82  & 41.21  & 43.23  & 44.53($\pm$1.00)  & 48.91($\pm$1.40) \\
    TAPD  & 50.72  & \textbf{46.58}  & 47.18  & \textbf{48.44}  & \textbf{47.68}  & \textbf{48.12}($\pm$2.84)  & \textbf{54.46}($\pm$2.63)   \\
    \midrule
    train data (300) \\
    Stancy &\textbf{61.87}  & 47.61  & 43.89  & 45.02  & 45.20  & 48.72($\pm$2.86)  & 53.28($\pm$2.70)\\
    BERT$_{SEP}$ & 52.80  & \textbf{52.55}  & 49.36  & 44.19  & 41.77  & 48.13($\pm$3.74)  & 53.30($\pm$2.68)    \\
    BERT$_{MEAN}$ &  56.45  & 51.42  & 46.98  & 43.27  & 43.59  & 48.34($\pm$3.85)  & 55.59($\pm$5.15)    \\
    BERT$_{TAN}$ &  52.19  & 50.05  & 44.09  & 43.49  & 49.65  & 47.89($\pm$3.85)  & 52.88($\pm$3.78) \\
    TAPD & 59.00  & 51.83  & \textbf{49.95}  & \textbf{55.64}  & \textbf{49.83}  & \textbf{53.25}($\pm$5.09)  & \textbf{60.17}($\pm$3.58) \\
    \midrule
    train data (450) \\
    Stancy & \textbf{61.95}  & 51.56  & 51.56  & 50.90  & 46.57  & 52.51($\pm$1.53)  & 58.04($\pm$2.57) \\
    BERT$_{SEP}$ &  54.35  & 53.63  & 51.35  & 50.59  & 41.02  & 50.19($\pm$3.09)  & 56.35($\pm$2.65)\\
    BERT$_{MEAN}$ &  56.49  & 44.67  & 47.82  & 48.98  & 48.53  & 49.30($\pm$4.04)  & 59.96($\pm$4.04)  \\
    BERT$_{TAN}$ &  53.92  & 44.21  & 47.25  & 48.43  & 48.26  & 48.41($\pm$3.46)  & 54.99($\pm$3.55)  \\
    TAPD & 59.63  & \textbf{60.88}  & \textbf{53.12}  & \textbf{59.88}  & \textbf{49.80}  & \textbf{56.66}($\pm$2.14)  & \textbf{63.18}($\pm$2.41)\\
    \bottomrule
    \end{tabular}}
\end{table}

\begin{table}
  \centering
  \caption{Few-shot stance detection performance on the UKP dataset.}
    \label{tab:fewshotukp}
    \resizebox{\linewidth}{!}{
    \begin{tabular}{ccccccccccc}
    \toprule
    {Model} & AB & CL & DP & GC & ML & MW & NE & SU & {$MacF_{avg}$} & {$MicF_{avg}$} \\
    \midrule
   train data (48) \\
    Stancy &  28.91  & 38.29  & 31.21  & 29.33  & 29.88  & 33.91  & 28.84  & \textbf{32.48}  & 31.61($\pm$1.14) & 32.09($\pm$1.29) \\
    BERT$_{SEP}$ & 27.82  & 36.86  & \textbf{34.37}  & 29.13  & 33.13  & 30.26  & 28.96  & 28.44  & 31.13($\pm$1.15) & 32.06($\pm$1.25)  \\
    BERT$_{MEAN}$ & 30.52  & 37.14  & 32.23  & 31.95  & 33.70  & 33.54  & 31.41  & 31.48  & 32.75($\pm$0.89) &33.49($\pm$1.23) \\
    BERT$_{TAN}$ &  29.09  & 35.97  & 31.44  & 33.00  & 32.47  & 32.15  & 31.67  & 30.15  & 31.99($\pm$1.48) &33.20($\pm$1.99) \\
    TAPD  & \textbf{33.38}  & \textbf{40.14}  & 32.90  & \textbf{34.19}  & \textbf{34.15}  & \textbf{35.16}  & \textbf{31.93}  & 27.79  & \textbf{33.71}($\pm$1.14) & \textbf{35.37}($\pm$1.36)\\
    \midrule
    train data (120) \\
    Stancy & 28.79  & 41.13  & 30.50  & 32.94  & 36.80  & 37.11  & 29.93  & 36.85  & 34.26($\pm$1.98) & 34.52($\pm$1.02) \\
    BERT$_{SEP}$ & 32.68  & 39.51  & 31.77  & 36.58  & 40.13  & 36.92  & 33.28  & 35.89  & 35.84($\pm$1.51) &36.49($\pm$1.36) \\
    BERT$_{MEAN}$ & 34.83  & 39.95  & 34.68  & 38.96  & 34.67  & 34.63  & 34.49  & 37.69  & 36.24($\pm$1.67) &37.51($\pm$1.21) \\
    BERT$_{TAN}$ & 36.17  & 41.53  & 35.08  & \textbf{39.54} & 37.22  & 36.76  & 35.97  & 38.13  & 37.55($\pm$1.68) & 38.54($\pm$2.08) \\
    TAPD & \textbf{36.57}  & \textbf{46.05}  & \textbf{36.14}  & 37.90  & \textbf{41.03}  & \textbf{41.20}  & \textbf{38.51}  & \textbf{39.57}  & \textbf{39.62}($\pm$1.73) &\textbf{40.05}($\pm$1.83) \\
    \midrule
    train data (240) \\
    Stancy & 32.65  & 46.72  & 37.22  & 33.93  & 43.49  & 45.00  & 37.80  & 41.52  & 39.79($\pm$1.68) & 40.01($\pm$1.45) \\
    BERT$_{SEP}$ & 36.00  & 45.74  & 36.27  & 38.46  & \textbf{44.82}  & 42.81  & 38.78  & 36.42  & 39.91($\pm$1.87) &40.06($\pm$1.93) \\
    BERT$_{MEAN}$ & \textbf{36.85}  & 47.05  & 36.49  & 39.13  & 39.87  & 41.42  & 39.30  & 39.48  & 39.96($\pm$1.29) & 40.52($\pm$1.00) \\
    BERT$_{TAN}$ & 34.74  & 48.92  & 38.88  & 35.59  & 44.25  & 42.37  & 42.35  & 41.57  & 41.08($\pm$2.53) &41.41($\pm$2.46)\\
    TAPD  & 36.65  & \textbf{51.87}  & \textbf{40.32}  & \textbf{39.89}  & 43.22  & \textbf{48.11}  & \textbf{42.75}  & \textbf{44.07}  & \textbf{43.45}($\pm$1.71)  & \textbf{43.46}($\pm$1.78)  \\
    \midrule
    train data (480) \\
    Stancy &33.93  & 51.93  & 41.43  & 36.29  & 44.54  & 47.82  & 45.38  & 45.93  & 43.41($\pm$1.61) &43.49($\pm$1.78)\\
    BERT$_{SEP}$ & 37.09  & 52.16  & 39.06  & 39.14  & \textbf{48.66}  & 46.77  & 44.54  & 44.52  & 43.99($\pm$1.02) &43.95($\pm$1.08)\\
    BERT$_{MEAN}$ &  37.84  & 53.47  & 39.76  & \textbf{41.49}  & 47.18  & 46.47  & 47.60  & 46.83  & 45.08($\pm$1.10) &45.13($\pm$1.09)\\
    BERT$_{TAN}$ & 35.35  & 56.54  & 42.58  & 37.49  & 48.29  & 47.34  & 47.65  & 47.08  & 45.33($\pm$1.37)  & 45.32($\pm$1.23)\\
    TAPD & \textbf{38.02}  & \textbf{57.23}  & \textbf{44.16}  & 41.13  & 46.64  & \textbf{49.50}  & \textbf{48.41}  & \textbf{50.70}  & \textbf{46.97}($\pm$1.63) &\textbf{47.06}($\pm$1.61)\\
    \midrule
    train data (720) \\
    Stancy & 34.83  & 56.08  & 40.01  & 36.42  & 47.13  & 49.36  & 47.99  & 45.43  & 44.66($\pm$0.83) & 44.44($\pm$0.83) \\
    BERT$_{SEP}$ & 37.79  & 53.94  & 39.62  & 39.14  & \textbf{51.13}  & \textbf{52.62}  & 47.50  & 47.72  & 46.18($\pm$0.54) &45.86($\pm$0.75)\\
    BERT$_{MEAN}$ & 38.57  & 56.97  & 41.95  & 39.53  & 49.10  & 51.42  & 50.14  & 48.41  & 47.01($\pm$1.05) &46.82($\pm$1.15) \\
    BERT$_{TAN}$ & 35.82  & \textbf{59.14}  & 42.77  & 38.49  & 49.15  & 50.45  & 49.21  & 50.08  & 46.89($\pm$0.64)  & 46.78($\pm$0.81) \\
    TAPD & \textbf{38.65}  & \textbf{57.22}  & \textbf{44.55}  & \textbf{41.80}  & 49.99  & 51.80  & \textbf{51.25}  & \textbf{53.27}  & \textbf{48.57}($\pm$1.07) &\textbf{48.55}($\pm$1.15) \\
    \bottomrule
    \end{tabular}}
\end{table}

\subsection{Analysis} \label{analysis}
In this section, we thoroughly analyze the design of our proposed model. First, we further test the performance of our model in cross target stance detection (described in Section \ref{crosstarget}). Then we analyze the impact of different components of our model (described in Section \ref{ablation}) and show the performance of the model under different hyperparameter choices (described in Section \ref{hyperparameter}). Moreover, we analyze target-aware verbalizer and conduct case study on the learned [\textit{mask}] token (described in Section \ref{mask}). Note that our analysis is conducted on the SemEval-2016 dataset.

\subsubsection{Cross-Target} \label{crosstarget}
Some previous works resort to cross-target learning to address few-shot stance detection, \textit{i.e.}, adapting out-of-target classifiers to a new target. However, different expressions and topics under various targets limit the transferability of learned models. Our TAPD achieves promising few-shot performance by better capturing the correlation between the target and the stance via prompts, showing great potential for cross-target stance detection. In this section, we investigate the cross-target performance of TAPD.
Following the previous work \cite{zhang2020enhancing}, three targets from the SemEval-2016 dataset, including Hillary Clinton (HC), Legalization of Abortion (LA), and Feminist Movement (FM) and two additional targets, \textit{i.e.}, Donald Trump (DT) and Trade Policy (TP),  are selected. Considering the correlation between targets, 8 cross-target stance detection tasks are constructed, including FM $\to$ LA, LA $\to$ FM, HC $\to$ DT, DT $\to$ HC, HC $\to$ TP, TP $\to$ HC, DT $\to$ TP and TP $\to$ DT. We compare with two baseline models, SKET \cite{zhang2020enhancing} and TPDG \cite{liang2021target}. SKET\cite{zhang2020enhancing} is a knowledge-based GCN model, which incorporated semantic-emotion knowledge into heterogeneous graph construction to bridge the gap between the source and destination target for cross-target stance detection. TPDG\cite{liang2021target} is the state-of-the-art method that leveraged in-target and cross-target graphs to derive the target-adaptive graph representation of the context for stance detection. 

\begin{table}
  \centering
  \caption{Performance comparison of cross-target stance detection in terms of $F1_{avg}$ on eight tasks.}
    \label{tab:crosstargetexperiment}
     \resizebox{\linewidth}{!}{
    \begin{tabular}{ccccccccc}
    \toprule
    Souce$\to$ Target & FM$\to$ LA & LA$\to$ FM & HC $\to$ DT & DT $\to$ HC & HC $\to$ TP & TP $\to$ HC & DT $\to$ TP & TP $\to$ DT  \\
    \midrule
     SKET \cite{zhang2020enhancing}  & 53.6  & 51.3  & 47.7  & 42.0  & 33.5  & 46.0  & 44.4  & 39.5 \\
     TPDG \cite{liang2021target} & 58.3  & \textbf{62.4}  & \textbf{50.4}  & 52.9  & \textbf{59.5}  & 49.8  & \textbf{51.2}  & 48.9 \\
    \midrule
     TAPD & \textbf{58.9}  & 58.8  & 49.5  & \textbf{54.2}  & 56.2  & \textbf{53.8}  & 45.2  & \textbf{50.8}  \\
    \bottomrule
    \end{tabular}}
\end{table}

Table~\ref{tab:crosstargetexperiment} shows the results over eight cross-target tasks. We can see that our model surpasses SKET model in all eight tasks. It achieves comparable result to TPDG, especially in FM $\to$ LA, DT $\to$ HC, TP $\to$ HC and TP $\to$ DT. This proves that our model also has the potential to solve cross-target stance detection by modeling the intrinsic correlation between the target and the stance.

\subsubsection{Ablation Experiment} \label{ablation}
In this section, we investigate the effectiveness of different components in our model. We design several variants of our model and present the results in Table~\ref{tab:ablationexperiment}. We first present the results of each single prompt as $P_1$, $P_2$, and $P_3$. Comparing the results of each single prompt and TAPD, we can tell that merging the results of multiple prompts can significantly boost the performance. However, the merging mechanism also matters. We also try to merge the results via a voting mechanism, \textit{i.e.}, choosing the majority label or the label generated by the best model ($P_3$), denoted as ``prompts vote'' in the table. Though achieving slight improvement, it still performs worse than TAPD, indicating the advantage of the distillation design. 
Moreover, we also investigate the effectiveness of self-distillation and denote its result as ``$P_3$-distillation''. It performs slightly better that $P_3$ but worse than TAPD, which also proves the necessity of introducing multiple prompts.

\begin{table}
  \centering
  \caption{Experimental results of ablation study.}
    \label{tab:ablationexperiment}
     \resizebox{\linewidth}{!}{
    \begin{tabular}{ccccccccc}
    \toprule
    {Model} & AT & CC & FM & HC & LA & {$MacF_{avg}$} & {$MicF_{avg}$} \\
    \midrule
    $P_1$ & 71.17  & 58.15  & 59.37  & 69.58  & 64.04  & 64.46  & 72.69 \\
    $P_2$ & 71.77  & 55.82  & 55.77  & 68.21  & 65.00  & 63.31  & 71.24 \\
    $P_3$ & 73.36  & 58.68  & 59.55  & 68.05  & 62.79  & 64.49  & 73.23\\
    \midrule
    $P_1$ with fixed verbalizer & 67.03  & 58.73  & 57.20  & 68.82  & 63.42  & 63.04  & 71.47 \\
    \midrule
    $P_3$-distillation & 73.05 & 58.79 & 61.12 &69.16 & 64.26 & 65.28 & 73.76 \\
    \midrule
    prompts vote & 72.20 & 52.01 & 60.45 & 67.72 & 66.53 & 63.78 & 73.64 \\
    TAPD  & \textbf{73.87}  & \textbf{59.32} & \textbf{63.93}  & \textbf{70.01} & \textbf{67.23} &  \textbf{66.87} & \textbf{74.80}  \\
    
    \bottomrule
    \end{tabular}}
\end{table}

To validate the effectiveness of target-aware verbalizer, we compare its performance with fixed verbalizer which is denoted as ``$P_1$ with fixed verbalizer''. We can see that $P_1$ performs consistently better with its fixed version, demonstrating the effectiveness of target-aware verbalizer.

\subsubsection{Hyper-parameter} \label{hyperparameter}

\begin{figure}
  \centering
  \includegraphics[width=0.8\linewidth]{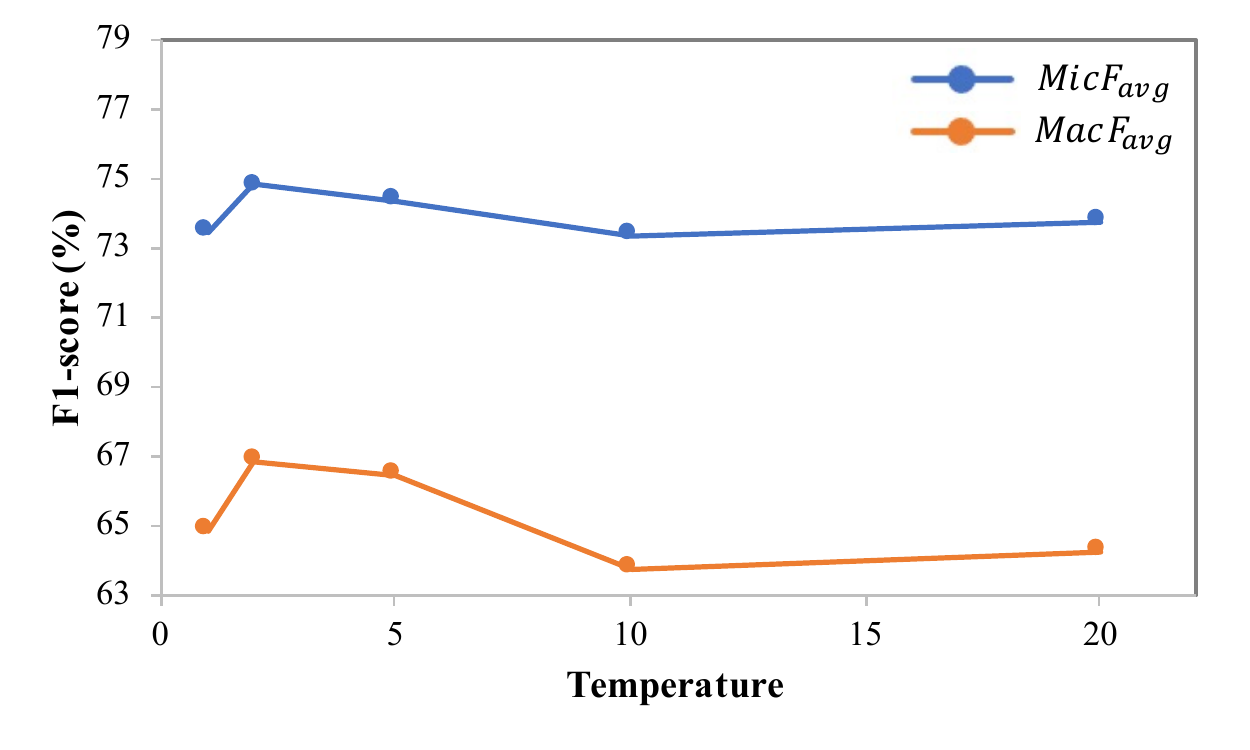}
  \caption{Performance of distillation at different temperatures.}
    \label{fig:temperature}
\end{figure}

The overall hyper-parameters of our model are selected via grid search over the entire space. Here we analyze the choice for two important hyper-parameters: the temperature and the balancing factor $\lambda$. To illustrate the impact of different temperatures, we show the performance of each temperature in {1, 2, 5, 10, 20} and fix the balancing factor to 0.8. The results are demonstrated in Figure~\ref{fig:temperature}. We can see that temperature $T=2$ works significantly better than higher or lower temperatures. When the temperature increases to more than 5, the performance tends to decline. We argue that using a higher value for temperature produces a softer probability distribution over classes and too soft probability distribution could not be suitable for learning information from the teacher model.

\begin{figure}
  \centering
  \includegraphics[width=0.8\linewidth]{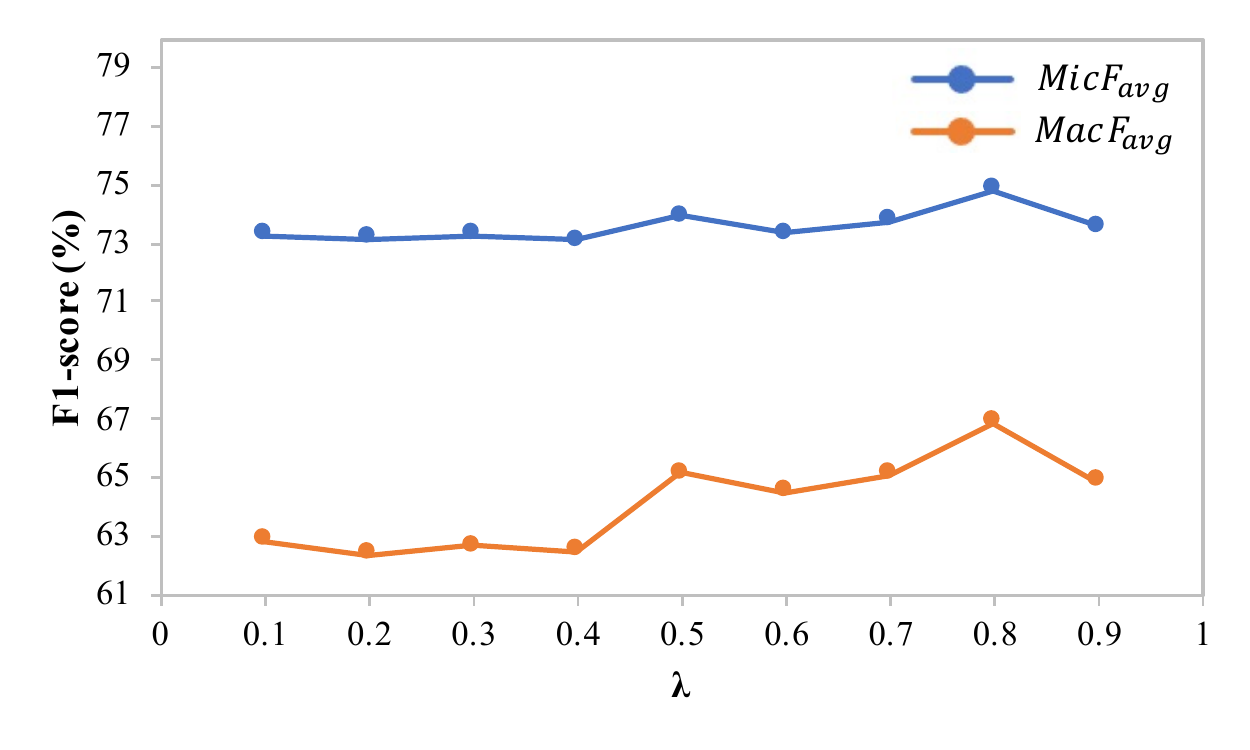}
  \caption{Performance of TAPD at different $\lambda$.}
    \label{fig:lamda}
\end{figure}

We also study the effects of different $\lambda$ and show the results in Figure~\ref{fig:lamda}. The temperature is set to 2 when showing the results. We can see that $\lambda=0.8$ works significantly better than higher or lower $\lambda$. 

\begin{table}
  \centering
  \caption{Top 10 closest words of target-aware stance vector in latent space.}
    \label{tab:diiferenttargets}
     \resizebox{\linewidth}{!}{
    \begin{tabular}{lp{3cm}p{3cm}p{3cm}}
    \toprule
    Target & Favor & None & Against   \\
    \midrule
    AT & science, social, public, reported, current, personal, patriotism, humanism, reason, bullying & stop, keep, maybe, nothing, behold, sorry, common, time, feel, anything  & lord, god, ministry, mercy, creation, darkness, false, christ, healing , ignorance \\
    \midrule
    CC & right, real, conservation, green, mankind, global, science, wwf, us, sustainability & doc, might, the, show, on, with, too, keep, common, from & denier, false, unnecessary, ship, engineering , official, works, child, further, short \\
    \midrule
    FM & inspired, empowered, women, equality, right, unique, inspiring, proud, fabulous, fun & here, present, mentioned, simple, different, inc, over, depicted, common, maybe & hated, fake, disgusting, fuck, false, oppression, violation, oppression, arrested, pathetic \\
    \midrule
    HC & campaign, proud, amazing, volunteered, right, loyal, great, so, honored, ready & third, too, live, here, 19, personal, voice, present, ft, people & fake, crazy, lies, trump, corruption, fraud, spy, bad, lying, fraud \\
    \midrule

    LA & optional, empowerment, rape, right, choice, feminist, hope, oppression, natural, needed & possible, over, basic, absent, hereditary, gradual, universal, maybe, rare, complete & life, death, murder, criminal, risk, infant, sacred, unnatural, wrong, creation \\
    \bottomrule
    \end{tabular}}
\end{table}

\subsubsection{Case study} \label{mask}
The target-aware verbalizer is designed to explore the correlation between the target and the stance. To validate the effectiveness of this module, we sample the top 10 crucial words that are closest to the target-aware stance vector in the latent space for all the labels across all the targets. Note that the top words are chosen from the vocabulary of the dataset. From Table~\ref{tab:diiferenttargets}, we can observe that our model can identify top words that are consistent with the topic and the stance label. Moreover, for the same stance label across different targets, our model can identify different top words for different targets, showing that our model is capable of capturing the correlation between the target and the stance.

\begin{table}
  \centering
  \caption{Case study.}
    \label{tab:casestudy}
     \resizebox{\linewidth}{!}{
  \begin{tabular}{p{73pt}p{200pt}p{30pt}p{30pt}p{30pt}}
    \toprule
    Target  &Text & Stance & Predicted Stance & [mask] word   \\
    \midrule
    Atheism & \#ILoveIslamBecause and v should love Islam because its a Deen instead of just a \#SemST  & Against & Against & false \\
    \midrule
    Feminist Movement & And girls just wanna have fun...damental rights \#SemST, &Favor & Favor & right \\
    \midrule
    Hillary Clinton & @HillaryClinton made me proud today!  Nothing like reinforcing what I already knew!  \#SemST & Favor &Favor & great \\
     \midrule
    Climate Change is a Real Concern & @neiltyson Rettet die Erde! Save the Earth! \#SemST &Favor & Favor & science\\
      \midrule
    Legalization of Abortion & `Snarck' - a snarky hack who puts down ordinary folk who dares question their pontifical commentaries  \#SemST & None & None & absent \\
    \bottomrule
    \end{tabular}}
\end{table}

We further study the word closest to the [\textit{mask}] token and present some cases for prompt $P_3$ in Table~\ref{tab:casestudy}. We can see that the predicted word are consistent with the stance label in most cases.

\section{Conclusion}
In this paper, we propose Target-Aware Prompt Distillation for stance detection. We design target-aware prompts and propose a novel verbalizer to better capture the correlation between the target and the stance. Besides, we propose a mutual-distillation mechanism to learn from diverse prompts for varying targets. Our model achieves state-of-the-art performance in both full-data and few-shot scenarios. In future work, we plan to investigate how to replace hand-crafted prompts via prompt-tuning techniques.

\section*{Acknowledgments}
This work is funded by the National Natural Science Foundation of China under Grant Nos. 62002347, U21B2046, and the National Key R\&D Program of China (2020AAA0105200). Huawei Shen is also supported by Beijing Academy of Artificial Intelligence (BAAI).

\bibliographystyle{ACM-Reference-Format}
\balance

\bibliography{sample-base}

\end{document}